\documentclass{article}

   \usepackage[final]{neurips_2019}

\usepackage[utf8]{inputenc} 
\usepackage[T1]{fontenc}    
\usepackage{hyperref}       
\usepackage{url}            
\usepackage{booktabs}       
\usepackage{amsfonts}       
\usepackage{nicefrac}       
\usepackage{microtype}      
\usepackage{times}
\usepackage{latexsym}
\usepackage{graphicx}
\usepackage[namelimits]{amsmath}
\usepackage{amssymb}
\usepackage{amsfonts}
\usepackage{multirow}

\title{HSCJN: A Holistic Semantic Constraint Joint Network for Diverse Response Generation}

\author{
  Yiru Wang \quad \quad Pengda Si \\
  Tsinghua Shenzhen International Graduate School\\
  Tsinghua University, P. R. China \\
  \texttt{\{wangyiru17,spd18\}@mails.tsinghua.edu.cn} \\
   \And
   Zeyang Lei \thanks{This work was done when Zeyang Lei was in Tsinghua University. } \\
    Baidu Inc.\\
    Beijing, P. R. China \\
   \texttt{zeyanglei@gmail.com} \\
   \And
   Guangxu Xun \\
   Department of Computer Science\\
   University of Virginia, USA\\ 
   \texttt{gx5bt@virginia.edu} \\
   \And
   Yujiu Yang \thanks{Corresponding author. This work was supported partially by Guangdong Basic and Applied Basic Research Foundation (No.2019A1515011387), and Shenzhen Basic Research Foundation (No.JCYJ201704is170118573).}\\
  Tsinghua Shenzhen International Graduate School\\
   Tsinghua University, Shenzhen, P. R. China \\
   \texttt{yang.yujiu@sz.tsinghua.edu.cn} \\
}

\begin{document}

\maketitle

\begin{abstract}
The sequence-to-sequence (Seq2Seq) model generates target words iteratively given the previously observed words during decoding process, which results in the loss of the holistic semantics in the target response and the complete semantic relationship between responses and dialogue histories. In this paper, we propose a generic diversity-promoting joint network, called Holistic Semantic Constraint Joint Network (HSCJN), enhancing the global sentence information, and then regularizing the objective function with penalizing the low entropy output. Our network introduces more target information to improve diversity, and captures direct semantic information to better constrain the relevance simultaneously. Moreover, the proposed method can be easily applied to any Seq2Seq structure. Extensive experiments on several dialogue corpuses show that our method effectively improves both semantic consistency and diversity of generated responses, and achieves better performance than other competitive methods.
\end{abstract}
\maketitle
\vspace{-0.35cm}
\section{Introduction}
\vspace{-0.15cm}
Recently, dialogue systems have attracted increasing attention in both academia and industry because of their potential applications and commercial values. Sequence-to-sequence (Seq2Seq) models form the cornerstone of popular response generation models \citep{SerbanSBCP16,SordoniGABJMNGD15,hangLL15}. However, neural dialogue systems based on Seq2Seq models tend to repeatedly generate universal and boring responses like ``I don't know.", ``Thank you.". Although being widely applied, conventional Maximum Likelihood Estimation (MLE) training could cause the low-diversity problem above \citep{LiMJ16}. Since high frequency words make up a big proportion of the training set, MLE encourages the model to excessively generate high frequency words.

Moreover, when training a Seq2Seq model traditionally, we iteratively maximize the log predictive likelihood of each true token in the target sequence, given the previously decoded tokens. Therefore, the model can only see the previous information during learning, unable to grasp the holistic information of the target sequence when decoding tokens. This also leads to the loss of complete semantic relationship between target sequences and source sequences.

As discussed, we argue that the current learning strategy heavily limits the Seq2Seq models to generate highly diverse responses, and the holistic semantic information of the target response as well as the global semantic relationship between responses and dialog histories are missing during generation process. Most previous solutions simply rely on external information or post-processing models to mask the deficiencies of the Seq2Seq model, while the problem of the Seq2Seq model itself has not been addressed. Therefore, in this paper, we hope to fully exploit the learning potential of the Seq2Seq models without any external information to improve diversity while better to constrain the semantic relevance of generated responses simultaneously. We propose a Holistic Semantic Constraint Joint Network (HSCJN) to predict the subsequent word set in each target utterance for direct supervision in decoding, which directly introduces more linguistic information from target utterances to increase diversity. More specifically, in the HSCJN, we require each hidden state in the decoder to predict the words in the target utterance which remain ungenerated, and the initial state of the decoder is required to predict all the words in the target utterance. Since the HSCJN enables the decoder network to see all words in the target utterance at every time step, our model also more likely captures direct semantic information such as keywords in target utterances to enhance relevance.

In this way, the relationship of representation spaces between source sequences and target sequences, and the transition between different decoder states could be better constrained. In addition, we consider that the entropy of the output distribution is low if the model is over-confident about high frequency words. Penalizing the low entropy output distribution can help regularize the model, optimize the predicted output distribution and alleviate the over-estimation of high frequency words. Therefore, we devise a maximum entropy based regularizer. Our learning framework can be used as a general joint training method with Seq2Seq models and requires no additional data or annotation. In general, our contributions are summarized as follows:
\vspace{-0.2cm}
\begin{itemize}
\item  We devise a joint training network to introduce future information in the decoding stage in open-domain dialogue generation, which can be applied to any Seq2Seq neural model. Our network introduces more linguistic information from target utterances to increase diversity, and likely captures key semantic information such as keywords in target utterances to enhance relevance in a direct manner.
\vspace{-0.15cm}
\item We regularize the model by penalizing low entropy output distribution at each time step in the decoder to alleviate the over-estimation of high frequency words, which also enables the loss function to consider every word in the vocabulary to improve diversity. 
\vspace{-0.15cm}
\item The experimental results on multi-turn dialogue datasets show the effectiveness of our method in terms of both diversity and relevance of generated responses.
\end{itemize}
\vspace{-0.4cm}
\section{Related Work}
\vspace{-0.2cm}
The diversity of generated responses is an important issue of common concern. \cite{SerbanSLCPCB17} and \cite{ZhaoZE17} proposed to introduce variational auto-encoders (VAEs) to Seq2Seq models to bring informativeness by increasing variability. Some researches proposed several beam search based approaches \citep{LiSSLCN17,SongYFZZZ18,VijayakumarCSSL16}. However, this kind of methods merely provide a criterion for reweighing response candidates, rather than producing more diverse responses in the first place. Besides, another previous works introduce additional information or knowledge such as the context \citep{SerbanSBCP16,TianYMSFZ17,YaoZFZY17} , keyword \citep{SerbanSBCP16,XingWWLHZM17,YaoZFZY17} or knowledge-base \citep{YoungCCZBH18,Ghazvinine18} into the response generation process to produce informative content. Although being effective, these approaches actually bypass the low-diversity problem by introducing the randomness of stochastic latent variables or additional information. The underlying Seq2Seq model remains sub-optimal in terms of diversity. \cite{LiGBGD16} proposed to use a Maximum Mutual Information (MMI) as an optimization objective to maximize the mutual information between messages and responses, but the MMI objective is used only during test time, and relies on many extra modules, like reverse models and beam search. \cite{ZhangNIPS18} proposed the Adversarial Information Maximization (AIM) model which considers explicitly maximizing mutual information during training to generate informative responses, but it still needs to train an extra backward model generating source from target, and implemented with complicated adversarial training strategy.

In other tasks, the word prediction technique has been applied in the neural machine translation \citep{WengHZDC17, LHostisGA16}. \cite{LinAAAI19} adds entropy to the loss function to make the sparse distribution more specifically concentrate on a small set of video segments in the VQA task. Unlike we consider entropy on the entire vocabulary, it only considers the entropy of audio, video, and words in the current sentence on the video segment. 
\begin{figure}
\centerline{\includegraphics[width=12cm,height=6.5cm]{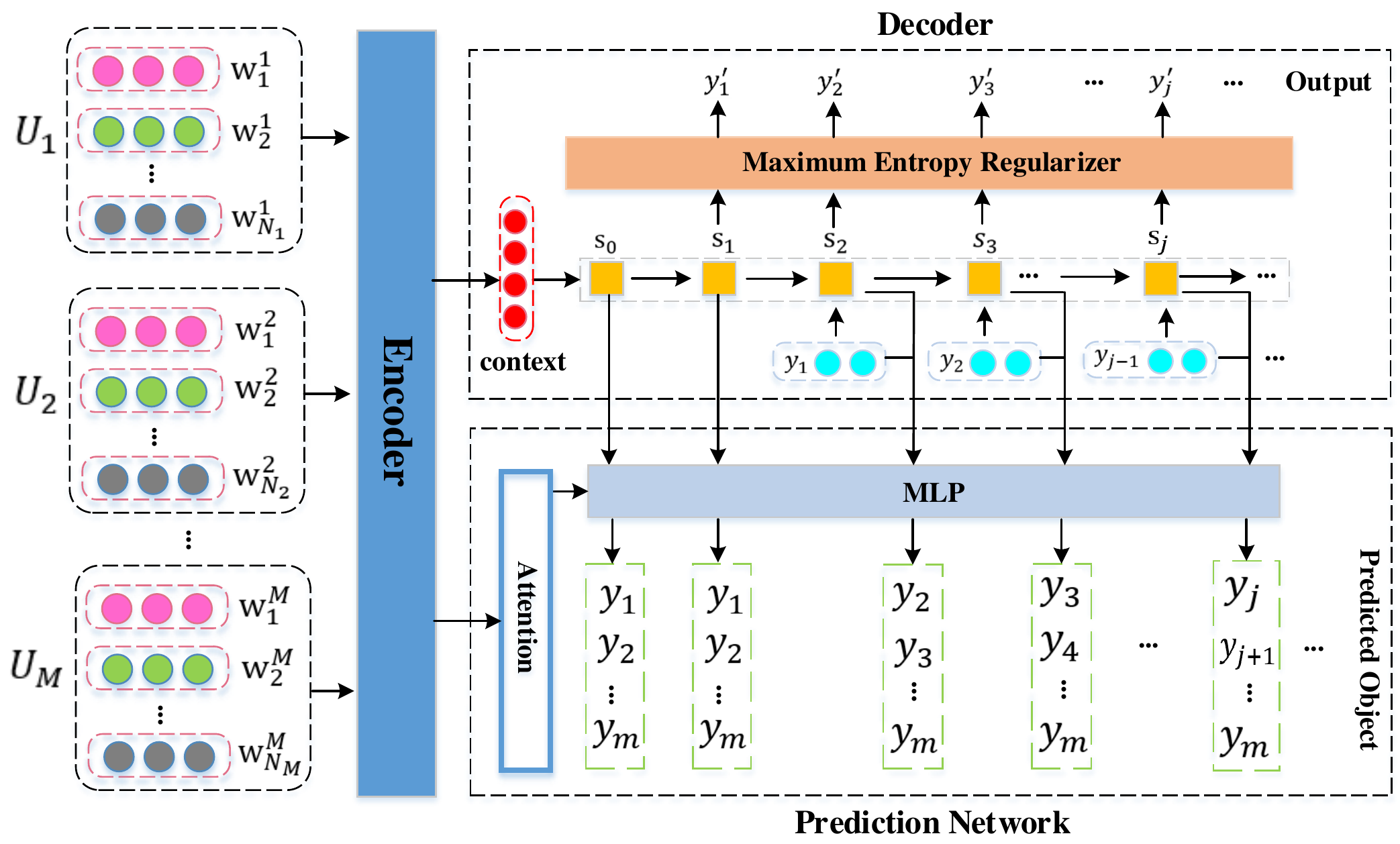}}
\caption{The architecture of the proposed HSCJN for multi-turn dialogue generation. The prediction network only works during the training phase which helps improve the Seq2Seq model for inference.}
\vspace{-0.2cm}
\label{fig:framework}
\end{figure}
\vspace{-0.2cm}
\section{HSCJN Model}
\vspace{-0.1cm}
\subsection{Task Definition}
Given a dialogue as a sequence of $M$ utterances $ \mathbb{U}=\{U_1,\ldots,U_M\} $, and each $U_i$ as a sequence of $N_i$ tokens $U_i = \{w_{i1},\ldots,w_{iN_i}\}$, where $w_{ij}$ represents the token at position $j$ in utterance $i$ from the vocabulary $V$, our task is to generate a response $Y =\{y_1,y_2,\ldots, y_m\}$ that is not only fluent and grammatical but also not repeated and trivial in content. Essentially, the goal is to estimate the conditional probability:
\begin{equation}
\begin{split}
  P(Y|\mathbb{U}) = \prod^m_{t=1}P(y_t|\mathbb{U}, y_{< t}), \text{ with } \mathbb{U} =  \left\{  [w_{1k}]_{k=1}^{N_1}, \ldots, [w_{Mk}]_{k=1}^{N_M} \right\}
\end{split}
\end{equation}
\subsection{An Overview of HSCJN}
Figure \ref{fig:framework} demonstrates the architecture of our model, in which we join a prediction network with the decoder network on each hidden state of the decoder. The encoder takes the word embedding sequences of context utterances as inputs, and obtains the hidden representations of the context. The decoder starts the generation of the target sequence from the initial state $s_0$. Since the initial state is responsible for the generation of the whole target sequence, we optimize the initial state by making prediction for all the target words to contain comprehensive target information. Similarly, at each time step in the decoder, we introduce the prediction network to predict the word set of the target subsequence that has not yet been generated.  The response is generated from the decoder, HSCJN applies a constraint network to each hidden state in the decoder to introduce more direct language information from the Seq2Seq model. There is a specific objective function for the HSCJN. In addition, we optimize the output distribution at each decoding step by adding a maximum entropy based regularizer to the final objective function to generate the predictive response.
\vspace{-0.1cm}
\subsection{Holistic Semantic Constraint Joint Network Design}
\vspace{-0.1cm}
In HSCJN, we require the hidden state at each time step in the decoder to predict the word set containing target words which remain ungenerated in the target utterance, where the order of words is not considered and we assume target words are independent with each other. In this way, at each time step, the decoder generates words not only conditioned on the previously generated subsequences within the original decoder network, but also under consideration of the future words not yet seen in the target sequence through our HSCJN. That is, our joint network HSCJN can introduce and utilize the global sentence information in target utterances for every token generation, beneficial for both diversity and relevance. Specifically, for each time step $j$ in the decoder, the hidden state $s_j$ is required to predict the word collection of $Y_{j\sim m} \triangleq(y_j, y_{j+1}, ... , y_m)$. The conditional probability $P_j$ of the prediction task in the HSCJN at hidden state $s_j$ is defined as follows:
\begin{gather}
\begin{split}
     P_j(Y_{j\sim m} |\Psi) =\prod\nolimits_{t=j}^mP(y_t|\Psi), \text{ with }P(y_t|\Psi ) = MLP([e(y_{j-1});s_j;c_j])
\end{split}
\end{gather}
where, $\Psi \triangleq \{y_{<j},U_1,\ldots,U_M\}$, and the set $Y_{j\sim m} $ is the word set of the future subsequence in target response $Y$ at time step $j$. $MLP$ is a multi-layer perceptron with two hidden layers using $tanh(.)$ as an activation function, followed by one output layer with $sigmoid(.)$ acting on each neuron. Here, we predict the target word set in a multi-classification way.  $e(y_{j-1})$ is the embedding of the word $y_{j-1}$, and $c_j$ is the context vector from the attention mechanism\citep{LuongPM15}.
\begin{gather}
\begin{split}
     c_j =\sum\nolimits_{i=1}^Ta_{ji}h_i, \text{ with }
   a_{ji}=\frac{exp(tanh(W_c[s_{j-1},h_i]))}{\sum_{k=1}^Texp(tanh(W_c[s_{j-1},h_k]))}
\end{split}
\end{gather}
where $W_c$ is weight parameters, $T$ is the input sequence length, $h_i$ is the hidden state of the encoder RNN at time step $i$, and $s_{j-1}$ is the hidden state of the decoder RNN at the previous time step ${j-1}$ .

Specifically, for the initial state $s_0$, HSCJN requires it to predict the word set containing all target words, so as to compress the overall information of the target sequence into the initial state. Therefore, the decoder can see the entire target sequence at the initial time step through the HSCJN. The conditional probability $P_0$ of the HSCJN at the initial state $s_0$  is defined as follows:
\begin{gather}
\begin{split}
  P_0(\widetilde{Y}| \mathbb{U}) = \prod_{t=1}^m P_0(y_t|\mathbb{U} ), \text{ with } P_0(y_t|\mathbb{U}) = MLP([s_0;c_0])
\end{split}
\end{gather}
where $\widetilde{Y}$ is the word set containing all target words in the target response $Y$, and $c_0$ is the context vector from the attention mechanism for the initial state $s_0$.

To optimize the HSCJN network, we add an extra likelihood function $L_{WP}$ into the training procedure:
\begin{equation}
  L_{WP}=-\frac{1}{m}\log P_0-\sum_{j=1}^m\frac{1}{m-j+1}\log P_j
\end{equation}
where $P_0$ and $P_j$ are as previously defined, the coefficient of the logarithm is used to calculate the average probability of each prediction. This loss function is used to guide HSCJN to accurately introduce the expected target semantic information.
\vspace{-0.2cm}
\subsection{Output Distribution Regularizer}
\vspace{-0.1cm}
When a dialogue model generates universal responses, the prediction of high frequency words is too confident, that is, the entire output probability distribution is concentrated on high frequency words. In result, the entropy of the output distribution is low. We consider that maximizing the entropy of the output distribution at each decoding step could help to regularize the model and produce more diverse responses. By this means, the token-level distribution $P(y_t|\mathbb{U}, y_{< t})$ is better constrained to relieve over-estimation of high frequency words. Therefore, we add a negative entropy to the negative log-likelihood loss function during training. To minimize the overall loss function, the model encourages the maximization of entropy. Specifically, the loss is expressed as:
\begin{gather}
\begin{split}
         L_{ME} =-\sum\nolimits_{t=1}^mH(p(y_t| \widetilde{\Psi})), \text{ with } H(p(y_t|\widetilde{\Psi})) = -\sum_{i=1}^{|V|}p(w_i|\widetilde{\Psi})\ast \log p(w_i|\widetilde{\Psi})
\end{split}
\end{gather}
Where  $\widetilde{\Psi} \triangleq \{y_{<t},U_1,\ldots,U_M\}$, $H(\cdot)$ is the entropy of the output distribution at decoding step $t$, $|V|$ is the length of the vocabulary $V$, and $w_i$ represents a word in the vocabulary.

This loss function not only penalizes the low entropy output distribution when predicting each token, but also considers the entropy over the entire vocabulary, so that the model likely takes into account more words to increase diversity.
\vspace{-0.2cm}
\subsection{Loss Function}
\vspace{-0.1cm}
We add $L_{WP}$ and $L_{ME}$ to the original negative log-likelihood loss function. The final loss function for model training is as follows:
\begin{equation}
  L=-\log P(Y|\mathbb{U})+\alpha L_{WP}+\beta L_{ME}
\end{equation}
where $\alpha$ and $\beta \in [0,1]$ are weight coefficients, which control the strength of the joint prediction task and the output distribution regularizer respectively. Our HSCJN builds a training objective at sentence level instead of the traditional token-level transition, considering the complete linguistic information in target utterances for every token generation.
\vspace{-0.25cm}
\section{Experiment}
\vspace{-0.1cm}
\subsection{Data Preparation}
We evaluate our proposed method on two multi-turn dialogue datasets, DailyDialog \citep{LiSSLCN17}  and OpenSubtitles \citep{Tiedemann12}. 

\textbf{DailyDialog:} It is a high-quality and less noisy dataset, which contains 13,118 multi-turn dialogues, separated into training/validation/test sets with 11,118/1,000/1,000 conversations. For the computational efficiency, we remove the dialogues with more than 300 tokens, which only makes up a small proportion of the whole dataset, and finally our training/validation/test sets of DailyDialog dataset contain 10,712 / 976 / 960 conversations respectively.

\textbf{OpenSubtitles:} It is a collection of movie subtitles. Following previous work \citep{Xu-1802-01345}, we treat each turn in the dataset as the target text and the two previous sentences as the source text. We randomly sample 200,000 / 50,000 / 10,000 dialogues for training, validation, and testing, respectively. Similarly, we also remove dialogues with more than 300 tokens, and finally our training/validation/test sets of OpenSubtitles dataset contain 199,992/ 49,995 / 9,984 dialogues respectively.
\vspace{-0.1cm}
\subsection{Baselines}
\vspace{-0.1cm}
\textbf{AttnSeq2Seq:} A vanilla Seq2Seq model with attention mechanism \citep{BahdanauCB14}. The encoder and decoder are both recurrent neural networks (RNN) with LSTM as the basic cell, and the encoder RNN is bidirectional.

\textbf{HRED:} HRED \citep{SerbanSBCP16} considers dialogue history in multi-turn dialogue generation at two levels: a sequence of words for each utterance and a sequence of utterances, and models this hierarchy of conversations accordingly.

\textbf{VHRED:} VHRED \citep{SerbanSLCPCB17} augments the HRED model with a stochastic latent variable at the decoder, trained by maximizing a variational lower-bound on the log-likelihood. The latent variable helps facilitate the generation of long utterances with more information content.

\begin{table} \small
\caption{\label{tab:01}The results of 1-turn response generation on DailyDialog and OpenSubtitles datasets}
\centering
\begin{tabular}{|c|c|c|c|c|c|c|c|}
  \hline
  \quad &\multicolumn{7}{|c|}{\textbf{DailyDialog Corpus}} \\
  \hline
  \textbf{Model} & \textbf{BLEU-1} & \textbf{BLEU-2} & \textbf{BLEU-3} & \textbf{BLEU-4}& \textbf{Distinct-1} & \textbf{Distinct-2} & \textbf{Distinct-3}\\
  \hline
  Attnseq2seq  &7.17 &0.15&0.03&0.02&0.031/247&0.092/523&0.162/754\\
  \hline
  HRED          &9.44 &1.85&0.83&0.36&0.049/276&0.145/516&0.252/651\\
  \hline
  VHRED         &11.28 &2.30&1.05&0.44&0.068/418&0.197/778&0.342/1007\\
  \hline
  \textbf{HSCJN} &11.05 &\textbf{2.60}&\textbf{1.29}&0.56&0.075/463&0.242/954&0.424/1235\\
  \hline
  \textbf{HSCJN(w/o ME)}&\textbf{11.48}&2.55&1.28&\textbf{0.69}&\textbf{0.084/536}&\textbf{0.276/1149}&\textbf{0.475/1481}\\
  \hline
  \textbf{HSCJN(w/o PN)}&10.57&2.30&1.13&0.58&0.063/386&0.200/788&0.357/1043\\
  \hline
  \quad & \multicolumn{7}{|c|}{\textbf{OpenSubtitles Corpus}}\\
  \hline
   Attnseq2seq  &13.27&0.88&0.28&0.12&0.006/530&0.022/1623&0.043/2741\\
  \hline
   HRED          &12.50&1.25&0.47&0.16&0.007/459&0.027/1391&0.050/2039\\
  \hline
   VHRED         &\textbf{14.64}&1.28&0.36&0.13&0.007/707&0.034/2627&0.073/4855\\
  \hline
  \textbf{HSCJN}     &13.29&1.45&\textbf{0.55}&\textbf{0.26}&\textbf{0.012/883}&\textbf{0.063/3550}&\textbf{0.131/6056}\\
  \hline
  \textbf{HSCJN(w/o ME)} &13.15&\textbf{1.48}&0.54&0.23&0.011/880&0.056/3411&0.114/5773\\
  \hline
  \textbf{HSCJN(w/o PN)} &12.61&1.44&0.51&0.25&0.008/621&0.039/2231&0.078/3650\\
\hline
\end{tabular}
\end{table}

\begin{table}
\caption{\label{tab:03}2-turns dialog generation results on DailyDialog Corpus}
\centering
\begin{tabular}{|c|c|c|c|c|c|c|c|}
  \hline
  \textbf{Model} & \textbf{BLEU-1} & \textbf{BLEU-2} & \textbf{BLEU-3} & \textbf{BLEU-4}& \textbf{Distinct-1} & \textbf{Distinct-2} & \textbf{Distinct-3}\\
  \hline
  HRED          &10.23 &1.78&0.72&0.33&0.037/437&0.107/827&0.192/1104\\
  \hline
  VHRED         &\textbf{12.24} &\textbf{2.26}&0.91&0.37&0.052/663&0.150/1265&0.269/1749\\
  \hline
  \textbf{HSCJN} &11.63 &2.25&\textbf{1.01}&\textbf{0.54}&\textbf{0.053/689}&\textbf{0.175/1500}&\textbf{0.313/2056}\\
  \hline
\end{tabular}
\end{table}
\vspace{-0.5cm}
\subsection{Model Settings}
\vspace{-0.1cm}
Our proposed method is generic since it can be combined with any Seq2Seq model. In our experiments, we use HRED as the basis of our learning framework. We initialize the recurrent parameter matrices as orthogonal matrices while all the bias vectors are set to $ \mathbf{0} $. Other parameters are initialized by sampling from the Gaussian stochastic distribution $\mathcal{N}(0,0.01)$. The vocabularies are limited to the most frequent 25K and 30K words for DailyDialog dataset and OpenSubtitles dataset respectively. We apply GRU with 500 hidden states and GRU with 1000 hidden states to  the encoders at word-level and utterance-level respectively, and LSTM with 500 hidden states to the decoder. The dimension of word embedding is set to 300. We use the Adam optimizer \citep{KingmaB14} to update the parameters, with a batch size of 8. The learning rate is 0.0002 and the dropout rate is 0.75. Meanwhile, we set $\alpha$ to 1 and $\beta$ to 0.13. For decoding during test time, we simply decode until the end-of-utterance symbol \emph{eou} occurs, using a beam search with a beam width of 5. All models in the baselines are implemented with the same settings.
\begin{figure}
    \centering
    \includegraphics[width=\textwidth,height =4.5cm]{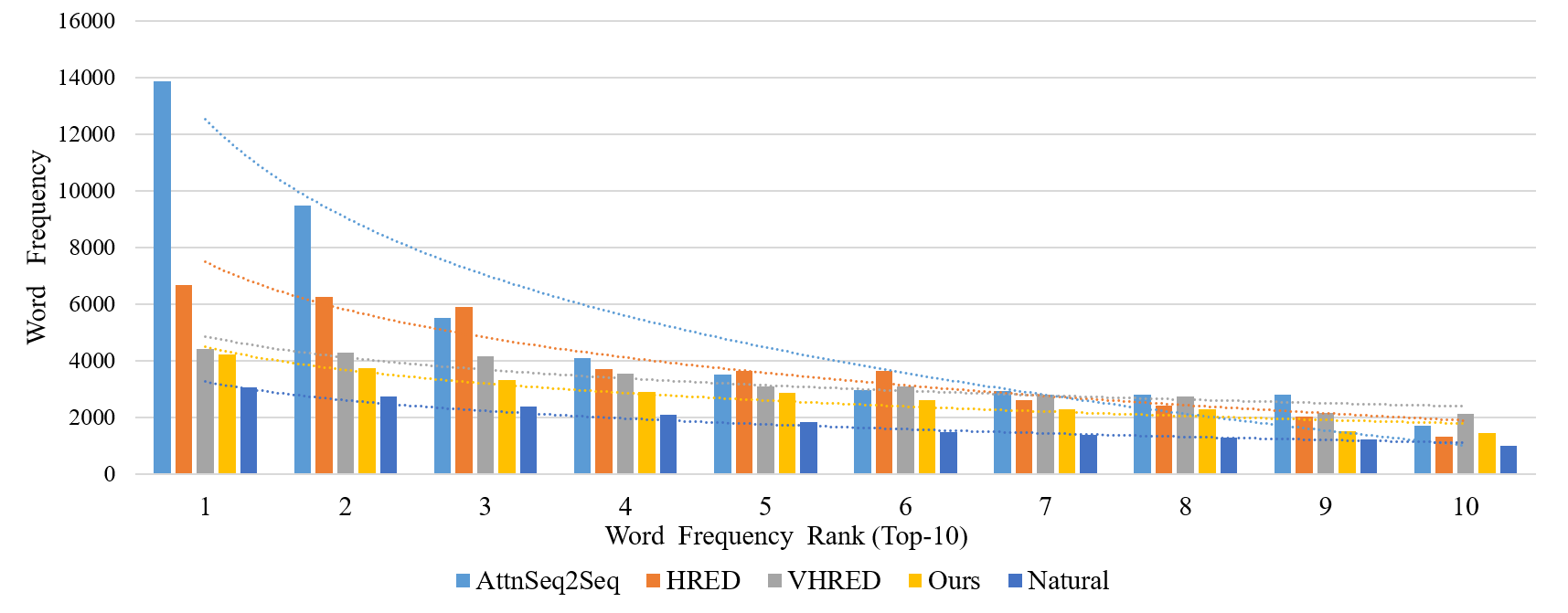}
    \caption{Word frequency distribution on generated responses on OpenSubtitles dataset. We draw the ten most frequent words, excluding punctuations.}
    \vspace{-0.3cm}
    \label{fig:wordDist}
\end{figure}
\vspace{-0.2cm}
\subsection{Automatic Evaluation}
\vspace{-0.2cm}
We adopt BLEU \citep{ChenC14,BLEU2002}, Distinct-1, Distinct-2 and Distinct-3 \citep{LiGBGD16} to evaluate the models at the quality and diversity level. The higher BLEU values demonstrate the responses are closer to the ground truth. Distinct-1, Distinct-2 and Distinct-3 are the proportion and number of distinct unigrams, bigrams, and trigrams in all the generated tokens, respectively. Higher Distinct-$n$ values are better for the overall diversity.

Table~\ref{tab:01} shows the experimental results of 1-turn response generation on DailyDialog corpus and OpenSubtitles corpus. It is obvious that our HSCJN generates remarkably more distinct unigrams, bigrams and trigrams than all the baselines on both two datasets. Besides, our model achieves the highest BLEU-2/3/4 values on both two datasets, compared with all the baseline models. Confirmed by our experiments, our model achieves an excellent performance in terms of both quality and diversity, regardless of the scale of datasets.

Furthermore, we conduct experiments on 2-turns dialogue generation. Given dialogue histories as input, we require models to generate the next two consecutive utterances. Since dialogues in the OpenSubtitles dataset contain only three turns, we conduct 2-turns response generation on the DailyDialog corpus. The results of 2-turns dialogue generation by our model and the other two multi-turn dialogue generation models are showed in Table~\ref{tab:03}. From the results, our model exceeds all baseline models with diversity improvement in multi-turn dialogue generation, and even higher BLEU-3 and BLEU-4.

To verify the effectiveness of our model in optimizing the output distribution, we perform word segmentation and word frequency statistics on the generated responses. Figure \ref{fig:wordDist} draws the distribution of ten most frequent words in responses generated by corresponding models on the OpenSubtitles dataset, excluding punctuations. ``Natural'' represents the natural distribution of the ground truth. The horizontal axis represents the rank of word frequency, and the vertical axis is word frequency values. The curves above columns fit the distribution trends of the word frequencies for different models, whose colors are consistent with the columns. It shows that the frequencies of words generated by our model are not as high as the baseline models, and the frequency distribution is flatter. Moreover, the distribution trends of our model and VHRED are basically consistent with the natural distribution, and our model is closer to the natural distribution in word frequency values than VHRED.
\vspace{-0.2cm}
\subsection{Ablation Study}
\vspace{-0.2cm}
We conduct the ablation study to examine the effectiveness of each mechanism, and the results are shown in Table~\ref{tab:01}. HSCJN(w/o ME) and HSCJN (w/o PN) represent the trained models without the maximum entropy regularizer and without the prediction sub-network respectively. On the DailyDialog dataset, HSCJN (w/o ME) generates the most distinct unigrams, bigrams and trigrams among all compared models, and also surpasses all the baseline models on Distinct-$n$ metrics on the OpenSubtitles dataset, indicating that our joint network can obviously generate more diverse responses. HSCJN(w/o ME) also achieves higher BLEU scores, which proves that our joint network can directly capture semantic information to enhance relevance. 

Since HSCJN(w/o PN) only adds a regularization item to the loss function in the training process of HRED, comparison with HRED model is sufficient for verifying the performance of the maximum entropy regularizer. From the results, HSCJN(w/o PN) achieves obvious improvement in both of the quality and diversity performance compared with HRED, demonstrating the effectiveness of the maximum entropy regularization. Both mechanisms contribute to the improvement of diversity and quality in response generation. The elimination of the prediction sub-network has greatest impact on the HSCJN model, indicating the importance of incorporating holistic semantic information. 
\vspace{-0.2cm}
\subsection{Manual Evaluation}
\vspace{-0.2cm}
Since automatic metrics for open-domain generative models may not be consistent with human perceptions, the quality scores from the human annotations are more reliable. Therefore, we further recruit human annotators to evaluate the quality of the generated responses. We randomly select 100 testing dialogues with responses generated by different models for each dataset and for both 1-turn and 2-turns generation. Responses generated by different models are randomly shuffled for each annotator. 5 annotators with linguistics experience are recruited to refer to the test dialogue histories and judge the quality of the responses of all compared models according to the following criteria:

\textbf{0:} The response cannot be used as a response to the conversation context. It is semantically unrelated or disfluent.

\textbf{+1:} The response can be used as a reply to the message, but it is too universal like ``Yes, I see.'', ``Thank you.'' and ``I don't know.''.

\textbf{+2:} The response is not only grammatical and relevant, but also informative and interesting.

Manual evaluation results for 1-turn response generation are presented in Table~\ref{tab:04}, which lists the percentage of each score and an overall average score. Among the three baselines, AttnSeq2Seq performs the worst and VHRED the best. Our model obtains a minimum of 0 points and a maximum of 2 points among all compared models on two datasets. This indicates that our model can generate fewer low-quality responses, as well as more semantically relevant and informative responses. The highest average score achieved by our model also confirms that our model outperforms the baselines.

\begin{table}
\caption{\label{tab:04} Manual evaluation results for 1-turn response generation}
\centering
\begin{tabular}{|c|c|c|c|c|c|c|c|c|}
  \hline
  \quad & \multicolumn{4}{|c|}{\textbf{Dailydialog Corpus}}& \multicolumn{4}{|c|}{\textbf{OpenSubtitles Corpus}}\\
  \hline
  \textbf{Score}&\textbf{0}&\textbf{+1}&\textbf{+2}&\textbf{Average}&\textbf{0}&\textbf{+1}&\textbf{+2}&\textbf{Average}\\
  \hline
  AttnSeq2Seq   &68.4\%&22.2\%&9.4\%&0.410&47.9\%&34.6\%&17.5\%&0.708\\
  \hline
  HRED          &55.2\%&30.7\%&14.2\%&0.590&28.6\%&50.0\%&21.4\%&0.928\\
  \hline
  VHRED         &47.7\%&30.0\%&22.3\%&0.747&29.4\%&39.8\%&30.8\%&1.014\\
  \hline
  \textbf{HSCJN} &\textbf{46.2\%}&28.8\%&\textbf{25.0\%} & \textbf{0.788}&\textbf{21.4\%}&45.2\%& \textbf{33.4\%}& \textbf{1.120}\\
  \hline
\end{tabular}
\end{table}
\vspace{-0.1cm}
\begin{table}\vspace{-0.2cm}
\caption{\label{tab:05}Manual evaluation results for 2-turns dialog generation on DailyDialog Corpus}
\begin{center}
\begin{tabular}{|c|c|c|c|c|}
\hline \textbf{Score} & \textbf{0} & \textbf{+1} & \textbf{+2} &\textbf{Average}\\ \hline
HRED           &54.0\%&24.4\%&21.6\%&0.676 \\
\hline
VHRED         &64.8\%&17.2\%&18.0\%&0.532 \\
\hline
\textbf{HSCJN} &\textbf{53.2\%}&22.4\%& \textbf{24.4\%}&\textbf{0.712} \\
\hline
\end{tabular}
\end{center}
\vspace{-0.2cm}
\end{table}

Table~\ref{tab:05} shows the manual evaluation results for 2-turns dialogue generation on DailyDialog dataset. In multi-turn generation, VHRED's performance is not good. Responses generated by VHRED may be informative but most of them are irrelevant to the context. Our model also outperforms the baselines in terms of relevance and informativeness, as well as the overall average score.
\vspace{-0.2cm}
\section{Case Study}
\vspace{-0.2cm}
\begin{table}
\caption{\label{tab:06}Examples of responses generated by different models given the multi-turn dialogue contexts}
\begin{center} \small
\begin{tabular}{|l|}
\hline
\textbf{Case 1: 1-turn response generation}\\ \hline
\textbf{Speaker A:} Good morning. Are you Mr.Liu? \\
\textbf{Speaker B:} My name is Liu Lichi. How do you do? \\
\qquad\qquad \textbf{$\cdots$} \\
\textbf{Speaker A:} Have you had any working experience? \\
\textbf{Speaker B:} Well, I worked at a supermarket during last summer holidays. \\
\textbf{Speaker A:} How are your English and computer skills? \\
\textbf{Speaker B:} I have passed the CET- 4 and 6. As far as computer is concerned, I can use the computer for\\
\qquad\qquad\quad  ~~  word processing. \\ \hline
\textbf{AttnSeq2Seq:} A great idea. \\
\textbf{HRED:} What do you do? \\
\textbf{VHRED:} I think so. \\
\textbf{HSCJN:} ~ That sounds great. How long have you been interested in the job?\\
\textbf{Human:}~ Okay. Mr.Liu, we'll inform you of the results within a week.\\ \hline\hline
\textbf{Case 2: 2-turns dialogue generation} \\ \hline
\textbf{Speaker A:} ~ Can I borrow this magazine from you? It's really interesting and I can't put it down.\\
\textbf{Speaker B:} ~ I am sorry, but I can't lend it to you now, for I haven't finished reading it. If you don't mind, \\
\qquad\qquad\quad  ~~ I can lend you some back numbers to you.\\ \hline
\textbf{HRED:} \emph {Speaker A:} ~ Thank you very much.  \\
\qquad \qquad \emph{Speaker B:} ~ You're welcome. \\
\textbf{VHRED:} \emph{Speaker A:} ~ Thank you very much.\\
\qquad \qquad \emph{Speaker B:} ~ Do you have any questions?\\
\textbf{HSCJN:} \emph{Speaker A:} ~ That's great.\\
\qquad \qquad \emph{Speaker B:} ~ It's too good as you like it. \\
\textbf{Human:} \emph{Speaker A:} ~ That would be very kind of you. By the way, is it a monthly magazine?  \\
\qquad \qquad \emph{Speaker B:} ~ No, it is a fortnightly. So, you see, I can get the new one quite soon.\\
\hline
\end{tabular}
\end{center}
\end{table}

Table~\ref{tab:06} presents the examples of 1-turn response and 2-turns dialogue generated by different models, given the multi-turn contexts between two speakers as inputs. ``Human'' lists the reference response in the dataset of the given context. We can see that Case 1 is an interview between an interviewer and a candidate. Our model captures that this is a job interview and produces a question matching the interview situation. In contrast, the responses generated by the baseline models are generic and irrelevant. In Case 2, our model captures the emotion information that speaker A likes the magazine, giving a more specific and informative response, and generates two consecutive turns matching different speakers' roles, while the results generated by baselines are universal and monotone. It can be found that our model generates obviously better responses with more specific details and higher diversity. Moreover, it also shows that our results are more relevant to the dialogue scenario.

\vspace{-0.2cm}
\section{Conclusion}
\vspace{-0.2cm}
In this paper, we investigate the low diversity issue in dialogue generation task. We propose a Holistic Semantic Constraint Joint Network to introduce future information into the decoding stage. In addition, we devise a maximum entropy regularizer into our loss function to penalize the over-estimation of high frequency words. In this way, the model can see the entire target sequence and consider all the words in the vocabulary in the learning process. It is worth mentioning that our model introduces the language information in the target sequences from the Seq2Seq model itself for diverse response generation, which does not depend on any external information and variables, also beneficial to capture holistic dialogue semantics to promote relevance. Moreover, our joint learning framework can be generalized to any end-to-end model. Extensive experiments show that our model produces more informative and relevant responses than several competitive baselines.


%
%
%
%
\bibliography{acl2019}
\bibliographystyle{acl_natbib}

\end{document}